\documentclass{article}

    \usepackage[preprint]{neurips_2024}

\usepackage[utf8]{inputenc} %
\usepackage[T1]{fontenc}    %
\usepackage{hyperref}       %
\usepackage{url}            %
\usepackage{booktabs}       %
\usepackage{amsfonts}       %
\usepackage{nicefrac}       %
\usepackage{microtype}      %
\usepackage{xcolor}         %
\usepackage{graphicx}
\usepackage{wrapfig} %
\usepackage{xcolor} %
\usepackage{colortbl} %
\usepackage{multirow}
\usepackage{booktabs}
\usepackage{caption}

\newif\ifdrafting
\draftingfalse %
\ifdrafting
    \newcommand{\lijie}[1]{\textcolor{purple}{[lf: #1]}}
    \newcommand{\tianhong}[1]{\textcolor{blue}{[tl: #1]}}
    \newcommand{\luming}[1]{\textcolor{red}{[lt: #1]}}
    \newcommand{\km}[1]{\textcolor{orange}{[km: #1]}}
    \newcommand{\ds}[1]{\textcolor{green}{[ds: #1]}}
    \newcommand{\chen}[1]{\textcolor{cyan}{[cs: #1]}}
    \newcommand{\radu}[1]{\textcolor{violet}{[rs: #1]}}

    \newcommand{\xuan}[1]{\textcolor{pink}{ #1}}

\else
    \newcommand{\lijie}[1]{}
    \newcommand{\ds}[1]{}
    \newcommand{\tianhong}[1]{}
    \newcommand{\luming}[1]{}
    \newcommand{\km}[1]{}
    \newcommand{\chen}[1]{}
    \newcommand{\radu}[1]{}

    \newcommand{\xuan}[1]{}
\fi

\newcommand{\name}{UniFluid}

\title{Unified Autoregressive Visual Generation and Understanding with Continuous Tokens}

\author{
Lijie Fan\textsuperscript{1,*} \hspace{.35em}
Luming Tang\textsuperscript{1,*} \hspace{.35em}
Siyang Qin\textsuperscript{1,*} \hspace{.35em}
Tianhong Li\textsuperscript{2} \hspace{.35em}
Xuan Yang\textsuperscript{1} \hspace{.35em}
Siyuan Qiao\textsuperscript{1} \hspace{.35em}
\\
\textbf{
Andreas Steiner\textsuperscript{1}
Chen Sun\textsuperscript{1}
Yuanzhen Li\textsuperscript{1}
Tao Zhu\textsuperscript{1}
Michael Rubinstein\textsuperscript{1}
Michalis Raptis\textsuperscript{1}
}
\\
\textbf{
Deqing Sun\textsuperscript{1,$^\dagger$} \hspace{.45em}
Radu Soricut\textsuperscript{1,$^\dagger$}
}
\\
\hspace{.1em} \textsuperscript{1}Google DeepMind \quad \textsuperscript{2}MIT \quad \textsuperscript{*,$^\dagger$} equal contribution
}

\begin{document}

\maketitle

\begin{figure}[h]
\centering
\vspace{-2em}
\includegraphics[width=0.98\textwidth]{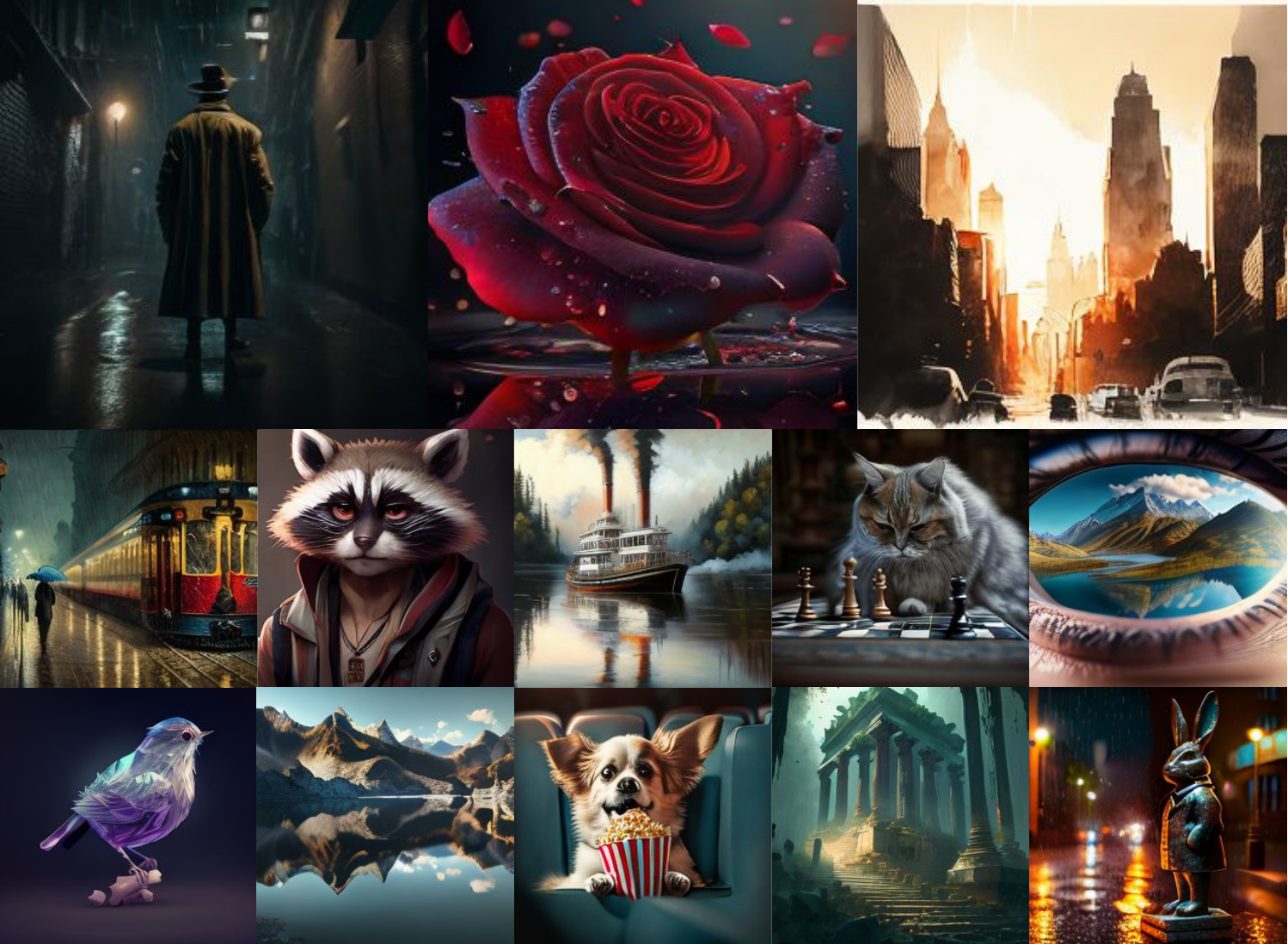}
\vspace{-2mm}
\caption{
\small Generated images from our \name~autoregressive model after aesthetic finetuning. 
}
\label{fig:teaser}
\end{figure}

\vspace{-2mm}
\begin{abstract}
\vspace{-2mm}
  We present \name, a unified autoregressive framework for joint visual generation and understanding leveraging continuous visual tokens. Our unified autoregressive architecture processes multimodal image and text inputs, generating discrete tokens for text and continuous tokens for image. 
We find though there is an inherent trade-off between the image generation and understanding task, a carefully tuned training recipe enables them to improve each other. By selecting an appropriate loss balance weight, the unified model achieves results comparable to or exceeding those of single-task baselines on both tasks.
  Furthermore, we demonstrate that employing stronger pre-trained LLMs and random-order generation during training is important to achieve high-fidelity image generation within this unified framework.
  Built upon the Gemma model series, \name~exhibits competitive performance across both image generation and understanding,
  demonstrating strong transferability to various downstream tasks, including image editing for generation, as well as visual captioning and question answering for understanding. 
 
\end{abstract}

\section{Introduction}

Large Language Models (LLMs) have recently advanced from text-centric architectures, such as BERT \cite{devlin2019bert} and GPT \cite{radford2018improving}, toward multimodal systems capable of understanding and generating content across different modalities. 
GPT-3~\cite{brown2020language} and PaLM~\cite{chowdhery2023palm} show that scaling language models leads to emergent capabilities, while Flamingo~\cite{alayrac2022flamingo} further demonstrates that incorporating visual inputs facilitates unified multimodal reasoning.
This trend toward unified vision-language model---using a single model for diverse tasks of visual understanding and generation---has promising potential for leveraging knowledge and reasoning abilities that transfer across different vision and language tasks, ultimately facilitating more robust and generalizable multimodal representation and modeling capabilities.

Motivated by the advantages and strong scaling properties of autoregressive models, 
coupled with their simplicity,
we investigate a pure autoregressive framework for unified visual generation and understanding, without the limitations introduced by vector quantization (VQ). 
In this paper, we introduce \name, a unified framework that leverages continuous visual tokens within an autoregressive architecture to jointly handle vision-language generation and understanding tasks. 
Building upon pre-trained Gemma \cite{team2024gemma2} on large-scale text corpus, \name~unlocks powerful visual generation and understanding capabilities through training with paired image-text data, and further allows these two tasks to mutually benefit each other within a single architecture.

Specifically, \name~adopts a unified autoregressive framework where both text and continuous visual inputs are embedded as tokens in the same space, enabling seamless joint training of image generation and understanding tasks. \name~integrates a continuous tokenizer~\cite{fan2024fluid,li2025autoregressive} for image generation and a pre-trained SigLIP~\cite{zhai2023sigmoid} image encoder for understanding tasks, while textual inputs are processed using a standard SentencePiece tokenizer~\cite{kudo2018sentencepiece}. The resulting multimodal sequences are modeled autoregressively using Gemma \cite{team2024gemma2} as the underlying transformer backbone. Task-specific prediction heads---a diffusion-based head for image generation and a cross-entropy head for text generation---ensure effective modality-specific training and inference, enabling \name~to efficiently learn shared representations that mutually enhance its generation and understanding capabilities.

Our experiments demonstrate several key advantages of the proposed unified training strategy. 
We find though there is a trade-off between the two tasks, a carefully tuned training recipe can allow the tasks to support each other and outperform the single-task baselines. 
Effectively balancing the loss between the tasks allows a single model that performs both with results superior to or on par with single-task models.
Moreover, the choice of pre-trained LLM backbone significantly impacts visual generation performance. We also find that while employing random generation order is essential for high-quality image synthesis, it is less critical for understanding tasks. Finally, our unified pre-trained models show strong generalization and transferability, achieving compelling results in downstream applications, including image editing and various vision-language understanding benchmarks.

\section{Related Works}

\textbf{Multimodal Large Language Models}.
Multimodal Large Language Models~\cite{alayrac2022flamingo,instructblip,lin2024vila,liu2023visual,liu2024improved,beyer2024paligemma,steiner2024paligemma} have shown significant performance in visual understanding tasks. Flamingo~\cite{alayrac2022flamingo} adopted a frozen LLM and vision encoder, utilizing perceiver with cross-attention to bridge the modalities. LLaVA~\cite{liu2023visual,liu2024improved} proposed instruction tuning over pre-trained LLMs with multimodal inputs to align a pre-trained image encoder into the LLM's embedding space, thereby enabling it with visual understanding and instruction following capabilities. 
MiniGPT-4 \cite{zhu2023minigpt} and mPLUG-Owl \cite{ye2023mplug} have shown vision encoders can be connected to LLMs through projection layers, demonstrating sophisticated visual reasoning capabilities.
The PaliGemma~\cite{beyer2024paligemma,steiner2024paligemma} series built upon the Gemma~\cite{team2024gemma,team2024gemma2} model family to develop versatile vision-language models capable of strong transfer to diverse downstream visual understanding tasks.

\textbf{Autoregressive Image Generation}.
While diffusion models~\cite{song2020score,rombach2021highresolution} have achieved impressive success in image generation, autoregressive image generation methods have also shown significant development, driven by their simplicity and 
closeness to 
LLM training paradigms.
A large body of research centers on tokenizing images into discrete tokens and applying autoregressive objectives to these discrete representations. Notable examples include Parti~\cite{yu2022scaling} and Muse~\cite{chang2023muse}. \cite{tian2024visual} proposes an approach that operates on image scales, progressively refining resolutions from coarse to fine through next-scale prediction. Alternatively, works such as MAR~\cite{li2025autoregressive}, Fluid~\cite{fan2024fluid}, and techniques employing per-token diffusion heads on top of LLM-predicted embeddings have explored autoregressive image generation with continuous visual tokens.

\textbf{Unified Multimodal Models}.
There is growing research interests in unifying visual generation and understanding within a single model. VQ-based models, such as Chameleon~\cite{team2024chameleon}, Emu~\cite{sun2023emu}, and Janus~\cite{wu2024janus,chen2025janus}, propose encoding visual inputs into discrete tokens and unifying tasks into next-token prediction within this discrete token space. Models with hybrid training targets, such as Transfusion~\cite{zhou2024transfusion,shi2024llamafusion} and Show-O~\cite{xie2024show}, aim to unify next-token prediction objectives with diffusion objectives within a single framework. MetaMorph~\cite{tong2024metamorph} maintains the autoregressive objective by regressing visual SigLIP~\cite{zhai2023sigmoid} features using an LLM, but necessitates a separate diffusion model to decode the predicted latent features into images.
Our approach distinguishes itself by performing per-token autoregressive generation using continuous visual tokens. This maintains the next-token prediction objective, while not being limited by the vector quantized tokens. 
\cite{sun2024multimodal} also explores using continuous tokens to generate multimodal outputs.
\section{Method}
In this section, we illustrate the architecture of our \name~model.
The model expects both image and text sequences as input and achieves joint training on both generation and understanding tasks, using next-token prediction as its training objective. 

\begin{figure}[t]
\centering
\includegraphics[width=\textwidth]{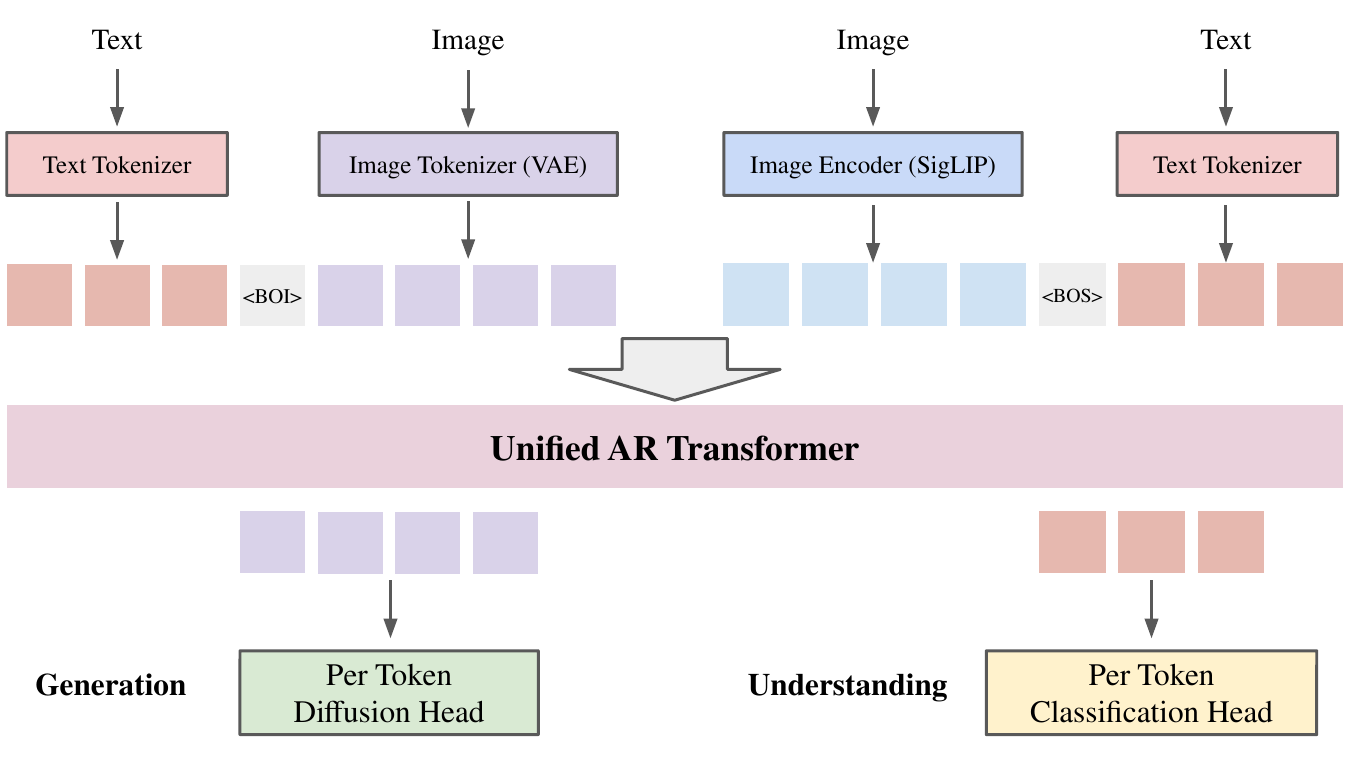}
\vspace{-5mm}
\caption{\small{High-level illustration of \name. \name~performs joint training of image generation and understanding tasks through next token prediction. %
For image embeddings, we use a VAE image tokenizer for generation, and a SigLIP image encoder for understanding. BOI/BOS stands for beginning of Image/Sentence.
}
}
\label{fig:model}
\vspace{-1mm}
\end{figure}

\subsection{Unified Autoregressive Modeling with Continuous Visual Tokens}

Our approach leverages the autoregressive paradigm to unify visual understanding and generation within a single framework. Given an ordered sequence of tokens $X = \{x^1, ..., x^n\}$, the autoregressive model
~\cite{parmar2018image,gregor2014deep,van2016conditional,van2016pixel}
factorizes the joint probability distribution as a product of conditional probabilities, effectively framing the generation task as a sequential ``next token prediction'' problem:
$p(X) = \prod_{i=1}^n p(x^i \mid x^1, ..., x^{i-1})$.
As shown in MAR~\cite{li2025autoregressive} and Fluid~\cite{fan2024fluid}, this autoregressive formulation is applicable for both discrete tokens and continuous tokens. In \name, we exploit this property to enable the generation of continuous visual tokens under the unified decoder-only framework. Our model is modality-agnostic; both text and image tokens are treated as elements within a long unified sequence, and their respective logits are predicted iteratively in an autoregressive manner by the backbone transformer.

To accommodate the distinct nature of text and image modalities, we employ modality-specific prediction heads to calculate the appropriate loss functions and sampling for each modality.
This unified approach allows the model to learn a shared representation space through the unified training procedure, facilitating synergistic learning and enabling seamless transitions between visual generation and understanding.

\subsection{Architecture}
As depicted in Figure~\ref{fig:model}, \name~employs a unified framework where both text and image inputs are tokenized and projected into a shared embedding space. This allows it to leverage a decoder-only transformer as the core backbone for the unified autoregressive task.
Text inputs are tokenized using the SentencePiece tokenizer. This results in discrete tokens with a vocabulary size of $V$.
For image generation, images are encoded into continuous visual tokens using a continuous variational autoencoder. To facilitate the process of image understanding, following PaliGemma, we used SigLIP as a separate image encoder to extract high-level information from the visual inputs.
\name~consists of a classification head to convert the transformer's text logits into a categorical distribution, and a diffusion head to convert image logits into a per-token probability distribution.

The inherent structure of text as a linear sequence aligns well with the standard 1D positional embeddings of the LLM, which are sufficient for text modeling and image understanding tasks. However, image tokens possess a 2D spatial structure. To capture this inherent 2D nature, we incorporate learnable 2D positional embeddings, which are added to the image token embeddings. %
Meanwhile, inspired by~\cite{yu2024randomized}, to achieve random order generation, a position embedding for the next predicted token is also added to each image token.
To enhance the model's ability to initiate and guide image generation, we prepend a "Beginning of Image" (BOI) token to the sequence of continuous image tokens. This BOI token serves as a distinct signal, indicating the start of the visual generation process. Given that the sequence length for generated image tokens is predefined (256 tokens for 256x256 images), an explicit "End of Image" token is unnecessary in our case.

\section{Implementation}
\subsection{Training}

\textbf{Per-token Classification Head for Discrete Text Tokens.}
We employ the same SentencePiece tokenizer as Gemma for text tokenization. The transformer's output logits for text are transformed into categorical probability distributions over the vocabulary, and we apply the standard cross-entropy loss, denoted as $L_{Text}$, to optimize the prediction of these discrete text tokens.

\textbf{Per-token Diffusion Head for Continuous Visual Tokens.}
We adopt the same continuous tokenizer as Fluid to embed 256x256 images into 32x32x4 continuous tokens, and use a patch size of 2 to merge 4 tokens into one. To model the per-token distribution of these predicted continuous visual tokens, we employ a lightweight MLP as a diffusion head. We adopt the same diffusion process and loss function, denoted as $\mathcal{L}_{Visual}$, as in~\cite{li2025autoregressive,fan2024fluid}, which is specifically tailored for continuous visual token prediction.
For the understanding task, the input image resolution is 224x224, and we use SigLIP as the image encoder. Note that the SigLIP features are only used as prefix for the understanding task during training, and no losses are added on top of them.

\textbf{Task-Specific Training Configurations.}

\textit{Image Understanding:} For image understanding tasks, the model is provided with image embeddings and question tokens as input prefix. Following PaliGemma, we apply a bidirectional attention mask to both image and question tokens. A causal attention mask is applied to the answer tokens, ensuring that the model only attends to previous answer tokens during autoregressive generation. The text token loss, $\mathcal{L}_{Text}$, is calculated specifically on the answer text tokens.

\textit{Image Generation:} Conversely, for image generation tasks, text prompts are provided as conditioning inputs. To maintain the appropriate information flow, we employ a bidirectional attention mask for the text prompt tokens, enabling them to attend to all other text tokens. A causal attention mask is applied to the image tokens, ensuring that each image token only attends to preceding image tokens. The visual token loss, $L_{Visual}$, is calculated on the generated image tokens.

\textbf{Unified Loss Function.}
The total training loss for \name~is a weighted sum of the text token prediction loss and the visual token prediction loss, defined as:
$\mathcal{L} = \mathcal{L}_{Visual} + \lambda\cdot \mathcal{L}_{Text}$
where $\lambda$ is a hyper-parameter that represents the weight assigned to the text token prediction loss, allowing us to balance the contributions of the two modalities during training.

\textbf{Training Details}.
We train the model with a batch size of 2048 using the AdamW optimizer with a learning rate of 1e-4. The training process consists of 1 million steps with a constant learning rate schedule and a warm-up period of 65k steps. Following~\cite{yu2024randomized}, for image generation, the image token order is randomly permutated during the initial 300k training iterations, then linearly anneals to raster between 300k and 600k iterations, and finally sticks to raster order for the subsequent 400k steps. Except for the comparison with Gemma-1, we use the Gemma-2 model series as the backbone transformer for all experiments in this paper.

\subsection{Inference}
For text decoding, we employ categorical sampling for each generated text prediction. The predicted token is then selected from the vocabulary $V$ based on the sampled probability distribution. 
We use the same decoding strategy as PaliGemma. Greedy decoding is used for all tasks except for downstream COCOcap (beam search n=2) and TextCaps (beam search n=3).
For image decoding, we use a diffusion sampling process to generate continuous visual tokens. with diffusion sampling step set to 100 in  our implementation.

As both text and image generation are performed at the token level, with predictions occurring one token at a time under a causal attention mechanism, we can efficiently utilize Key-Value (KV) caching. This optimization is applicable to both discrete text tokens and continuous visual tokens, significantly accelerating the inference process.

\section{Experiment}

\subsection{Setup}

\textbf{Datasets}.
We train our models using the WebLI dataset~\cite{chen2022pali}, a collection of high-quality image-text pairs.
For visual generation, we follow Fluid to employ a WebLI subset of image and text descriptions specifically for the generation task.
For visual understanding, consistent with PaliGemma, we leverage the image-text description pairs and image question-answer pairs that are also available within WebLI.

\textbf{Evaluation Metrics}.
We assess the image generation quality using the FID~\cite{heusel2017gans} score on 30K images of the MS-COCO~\cite{lin2014microsoft} training set and evaluate performance on the GenEval~\cite{ghosh2023geneval} benchmark, where we use the original text prompt without any rewrites. %
For evaluating visual understanding performance, we use the caption CIDEr score on MS-COCO. Given our similar training dataset and setup to PaliGemma, we also evaluate the finetuning performance on a variety of captioning and question answering tasks. We report the average score on 4 Captioning tasks, including COCOcap~\cite{lin2014microsoft}, Screen2Words~\cite{wang2021screen2words}, TextCaps~\cite{sidorov2020textcaps}, WidgetCap~\cite{li2020widget}, and 20 QA tasks, including OKVQA~\cite{marino2019ok}, AOKVQA-MC~\cite{schwenk2022okvqa}, AOKVQA-DA~\cite{schwenk2022okvqa}, GQA~\cite{hudson2019gqa}, NLVR2~\cite{suhr2018corpus}, AI2D~\cite{kembhavi2016diagram}, ScienceQA~\cite{lu2022learn}, RSVQA-lr~\cite{lobry2020rsvqa}, RSVQA-hr (test/test2)~\cite{lobry2020rsvqa}, ChartQA (human/aug)~\cite{masry2022chartqa}, VizWizVQA~\cite{gurari2018vizwiz}, TallyQA (simple/complex)~\cite{acharya2019tallyqa}, CountBenchQA~\cite{beyer2024paligemma}, TextVQA~\cite{singh2019towards}, DocVQA~\cite{mathew2021docvqa}, InfoVQA~\cite{mathew2022infographicvqa}, ST-VQA~\cite{biten2019scene}.

In the following sections, we present the experimental results  obtained under different configurations of \name, providing insights into the relationship between the two tasks and highlighting key design choices for \name~training.

\subsection{Main Results}

\textbf{Unified Training Improves Generation Performance.}
To evaluate the effectiveness of the unified training framework and determine whether unified training offers advantages compared to training separate models for different tasks, we perform controlled experiments to analyze the performance of models trained with a single task.

\begin{table}

\caption{\small Unified training achieves better generation performance than text-to-image only training. We evaluate the performance using MS-COCO zero-shot FID and GenEval score.}
\label{tab:unified_t2i}
\vspace{2mm}
\centering
\resizebox{0.46\textwidth}{!}{%
\begin{tabular}{c|c|cc}
\toprule
\textbf{Training Target} & \textbf{Size} & \textbf{FID} $\downarrow$  & \textbf{GenEval} $\uparrow$ \\
\midrule
T2I only & 0.7B & 9.71 & 0.50 \\
Unified & 0.7B & \bf 8.39 & \bf 0.52 \\
\midrule
T2I only & 2B & 7.88 & \bf 0.59 \\
Unified & 2B & \bf 7.20 & \bf 0.59\\
\bottomrule
\end{tabular}
}
\end{table}
We first compare the visual generation performance of the model trained under the unified training objective with the performance of a text-to-image model (T2I only), trained solely with the visual autoregressive objective for the generation task. We ensure that the total number of visual tokens for training is the same for the visual generation loss in both the unified model training and text-to-image only training scenarios.
The generation performance comparison is presented in Table~\ref{tab:unified_t2i}. The unified model achieves better performance compared to the T2I only model, despite both models having observed the same number of tokens for the visual generation task. This suggests that unified model training can be beneficial for visual generation tasks, and that visual understanding ability has the potential to unlock enhanced visual generation quality.

\textbf{Trade-off Between Generation and Understanding.}
We also investigate whether the visual generation task can contribute to improved visual understanding performance. 
In the \name~unified training setup, the hyperparameter $\lambda$ controls the balance between the losses applied to image tokens and text tokens.

\begin{figure}[p]
    \centering
    \begin{minipage}{\textwidth} %
        \centering
        
\hspace*{-19mm}
        \includegraphics[width=1.25\textwidth]{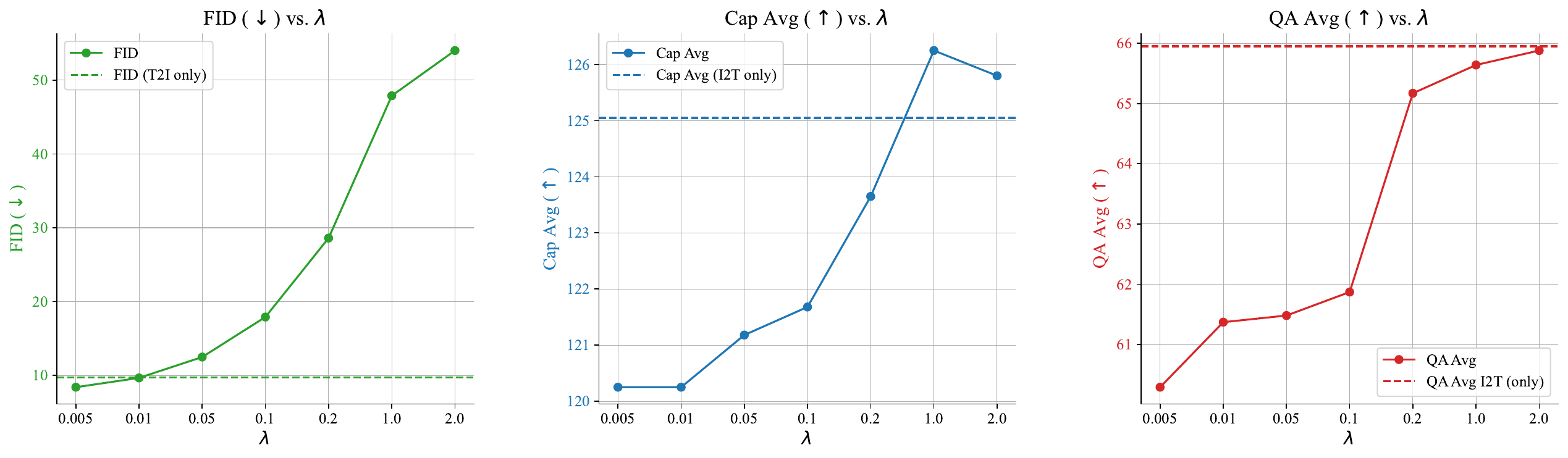}
        \captionof{figure}{\small Plot of image understanding and generation performance with varying $\lambda$ during training. We show the FID on MS-COCO, the average score for downstream captioning tasks (Cap Avg), and the average score for downstream QA tasks (QA Avg) with different $\lambda$ on the three figures. Green dashed lines shows the FID of T2I-only model, blue and red dashed lines shows the downstream captioning average and QA average for I2T-only model, respectively. In practice, smaller $\lambda$ has better trade-off between two tasks. }
        \label{fig:und}

                \vspace{2mm}

\captionof{table}{\small Image generation and understanding results with different $\lambda$.
$\lambda=0.1$ roughly makes the loss for generation and understanding of the same scale.
We present MS-COCO FID and the average captioning and QA results for downstream understanding tasks, compared to the image-to-text (I2T) only baseline. A smaller $\lambda$ like 0.005
is recommended in most cases as it preserves most of the image understanding capability while enabling the generation of high-quality images and outperforms the text-to-image (T2I) only baseline.
}
\label{tab:lambda_exp}
\resizebox{\textwidth}{!}{%
\begin{tabular}{c|c|c|ccccccc}
\toprule
\multirow{2}{*}{\textbf{Task}} & \multirow{2}{*}{\textbf{T2I only}} & \multirow{2}{*}{\textbf{I2T only}} & \multicolumn{7}{c}{ \textbf{Unified, $\lambda$}} \\
& & & \textbf{0.005} & {0.01} & {0.05} & {0.1} & {0.2} & {1.0}  & {2.0}\\
\midrule
Generation (FID)  $\downarrow$       &  9.71  & - & \bf 8.39 & 9.65 & 12.48 & 17.90 & 28.60 & 47.89 & 54.02
\\
Understanding (Cap Avg) $\uparrow$        & -   & 125.05 & 120.25 & 120.25 & 121.18 & 121.68 & 123.65 & \bf 126.25 & 125.80 \\
Understanding (QA Avg) $\uparrow$        & -   & \bf 65.95 & 60.29 & 61.37 & 61.48 & 61.87 & 65.17 & 65.64 & 65.88 \\

\bottomrule
\end{tabular}
}

                \vspace{10mm}

\hspace*{-14mm}
\includegraphics[width=1.2\textwidth]{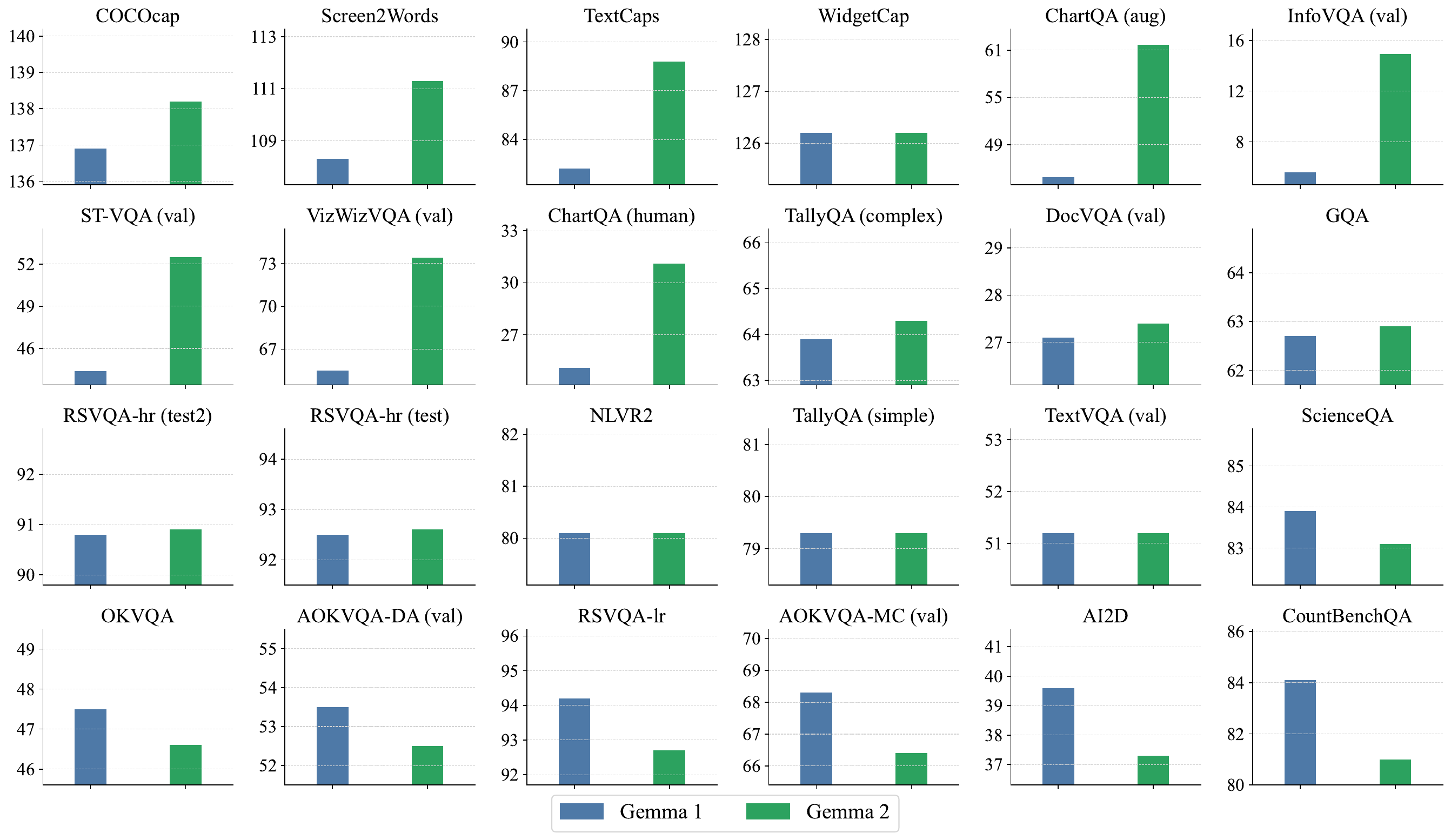}
        \captionof{figure}{\small Downstream image understanding performance (after finetuning) on each benchmark (higher is better). Blue bars shows performance of model backbone initialized with Gemma-1 2B and green bars shows model backbone initialized with Gemma-2 2B. Initializing with stronger LLM help improve \name~visual understanding performance on most datasets.
        }
        \label{fig:ds_gemma}
    \end{minipage}
\end{figure}

In Table~\ref{tab:lambda_exp} and Figure~\ref{fig:und}, we present the understanding and generation results with varying $\lambda$ of the 0.7B model. We compare the transfer performance to downstream understanding tasks between the unified model with different $\lambda$ and a image-to-text model (I2T only), trained solely with the image understanding objective.  
Within the unified training setup, a trade-off exists between visual generation and understanding tasks, which can be effectively controlled by adjusting the loss mixing weight, $\lambda$. While increasing $\lambda$ can improve image understanding performance, ultimately exceeding the I2T-only baseline for downstream captioning, it concurrently diminishes the image generation capabilities.

\begin{figure}[p]
\vspace*{-15mm}
\hspace*{-16mm}
\centering
\includegraphics[width=1.22\textwidth]{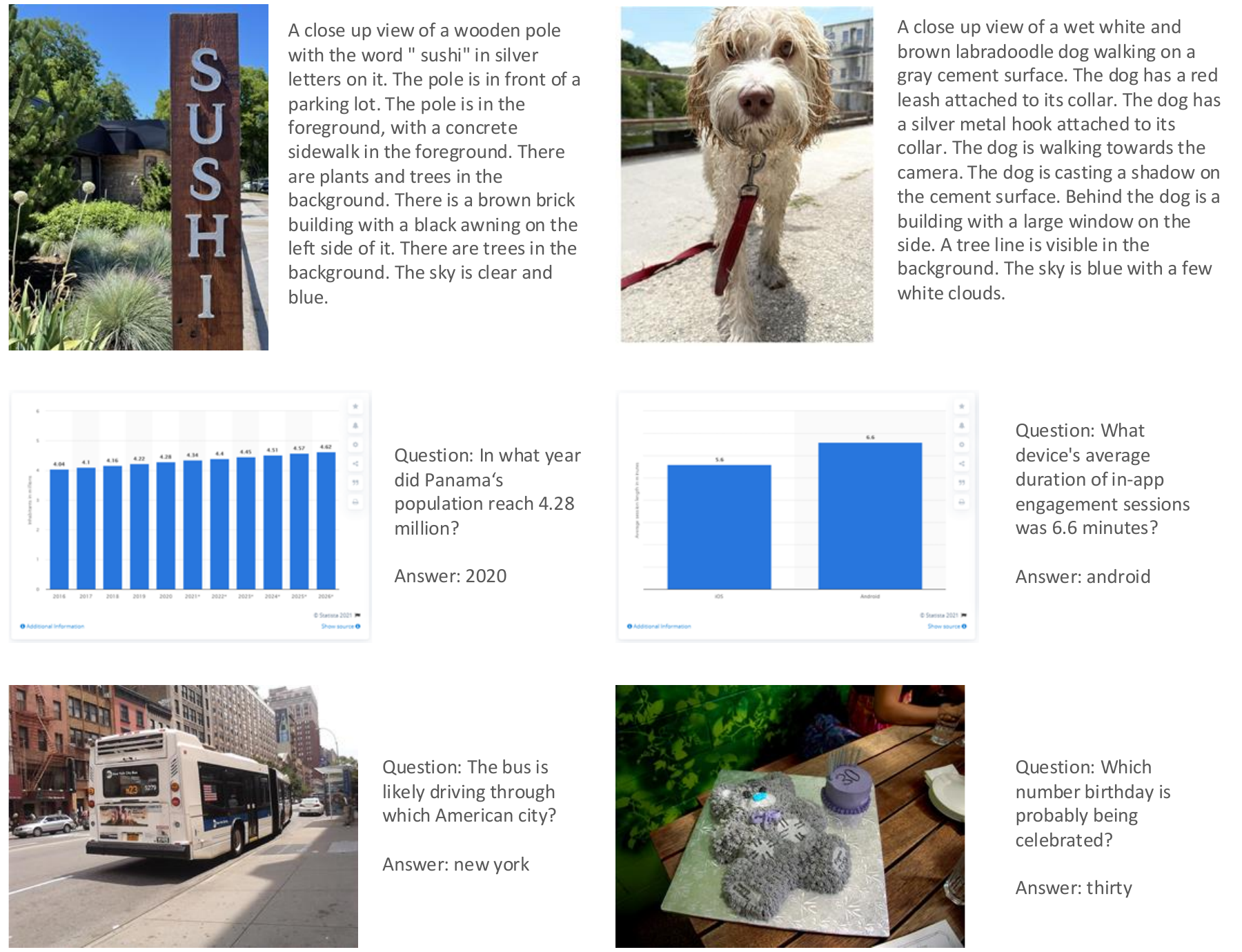}
\vspace*{-5mm}
\captionof{figure}{
\small Finetuned \name{} model demonstrates strong image-to-text capability on image captioning and question answering.
}
\label{fig:i2t}

\vspace*{3mm}
\hspace*{-16mm}
\centering
\includegraphics[width=1.22\textwidth]{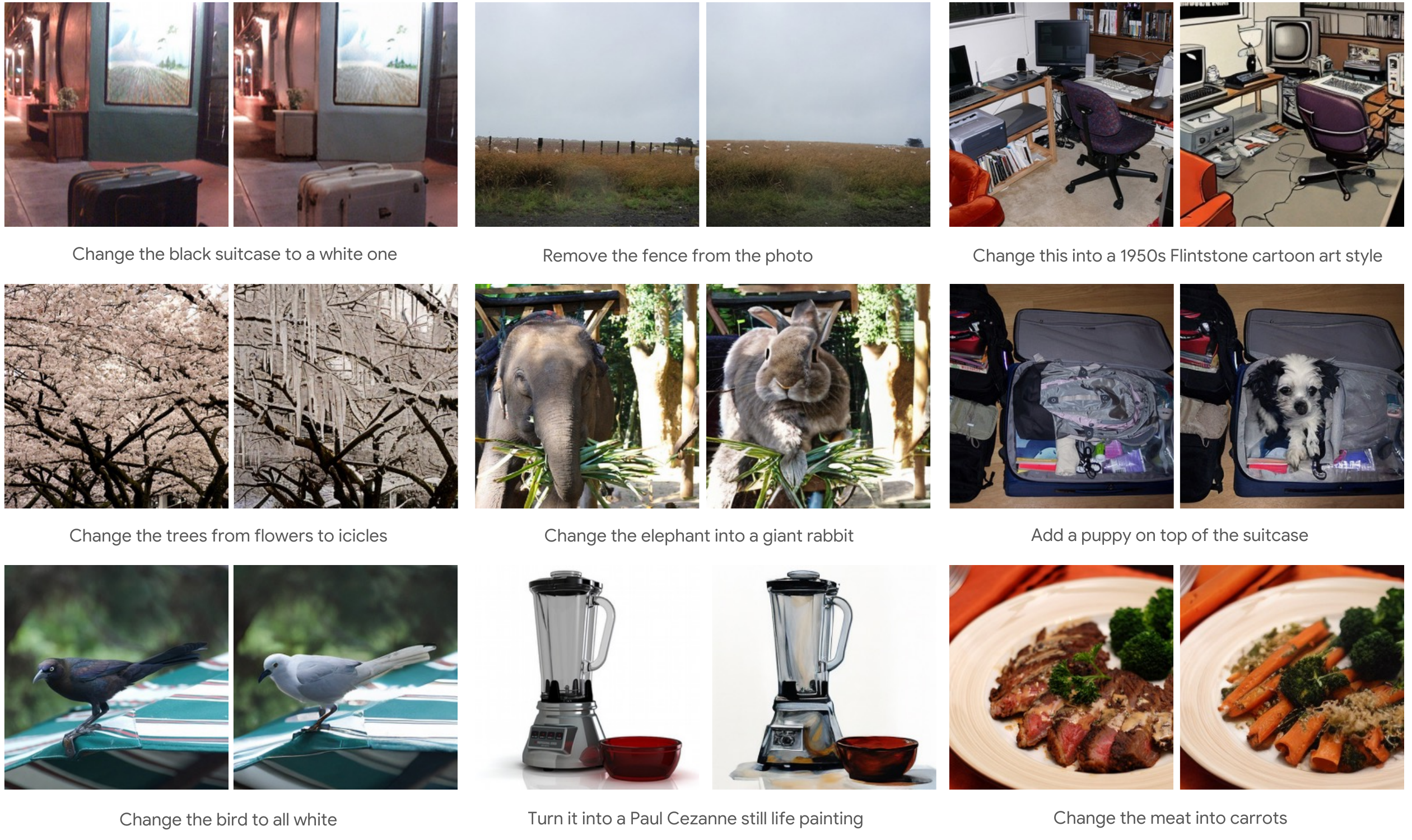}
\vspace*{-7mm}
\captionof{figure}{
\small Image editing results on evaluation benchmark from \name~autoregressive model after fine-tuning. It handles multiple editing tasks effectively, including object removal, insertion, style and color changes.
}
\label{fig:editing}

\end{figure}

In most scenarios, a smaller $\lambda$ value (e.g., 0.005) is advisable, maintaining a significant proportion (over 90\%) of image understanding while supporting the generation of high-fidelity images. Larger $\lambda$ values, in contrast, strongly favor image understanding but result in a rapid drop of image generation ability, as indicated by a sharp rise in FID score when $\lambda$ exceeds 0.1. Qualitative results for image captioning and question answering, demonstrating the understanding capabilities of the fine-tuned model based on the unified model with Gemma-2 2B as backbone LLM and $\lambda=0.005$ are presented in Figure~\ref{fig:i2t}.

\textbf{Better Pre-trained LLM Backbone Leads to Better Visual Generation and Understanding.}
We investigate the effect of pre-trained LLMs within the unified model training setup, specifically examining whether more powerful LLMs contribute to enhanced image understanding performance and superior visual generation quality. To this end, we conducted experiments using Gemma-1 2B~\cite{team2024gemma} and Gemma-2 2B~\cite{team2024gemma} as backbone LLMs. 
Gemma-2 is a stronger LLM than Gemma-1 with 10\% average improvements across different text benchmarks.

\begin{table}[h]
\caption{\small Performance comparison of image generation and understanding of \name~trained with different LLM backbone. FID and CIDEr are measured on MS-COCO. Gemma-2 gets much better performance compared to Gemma-1, for both image understanding and generation tasks.}
\label{tab:backbone_llm}
\centering
\resizebox{0.82\textwidth}{!}{
\begin{tabular}{ccc|ccc}
\toprule
\multirow{2}{*}{Pretrained Model} & \multicolumn{2}{c|}{Generation} & \multicolumn{3}{c}{Understanding} \\ \cmidrule{2-6} 
                 & COCO FID $\downarrow$  & GenEval $\uparrow$  & COCO CIDEr $\uparrow$ &  Cap Avg $\uparrow$ &  QA Avg $\uparrow$ \\ 
                 \midrule
Gemma-1          & 9.73     & 0.52    & 38.02 & 113.40 & 60.21 \\
Gemma-2          & \bf 7.20      & \bf 0.59    & \bf 40.91 & \bf 116.13 & \bf 62.10 \\ 
\bottomrule
\end{tabular}
}

\end{table}

\begin{figure}[p]
\vspace*{-22mm}
\hspace*{-22mm}
\centering
\includegraphics[width=1.3\textwidth]{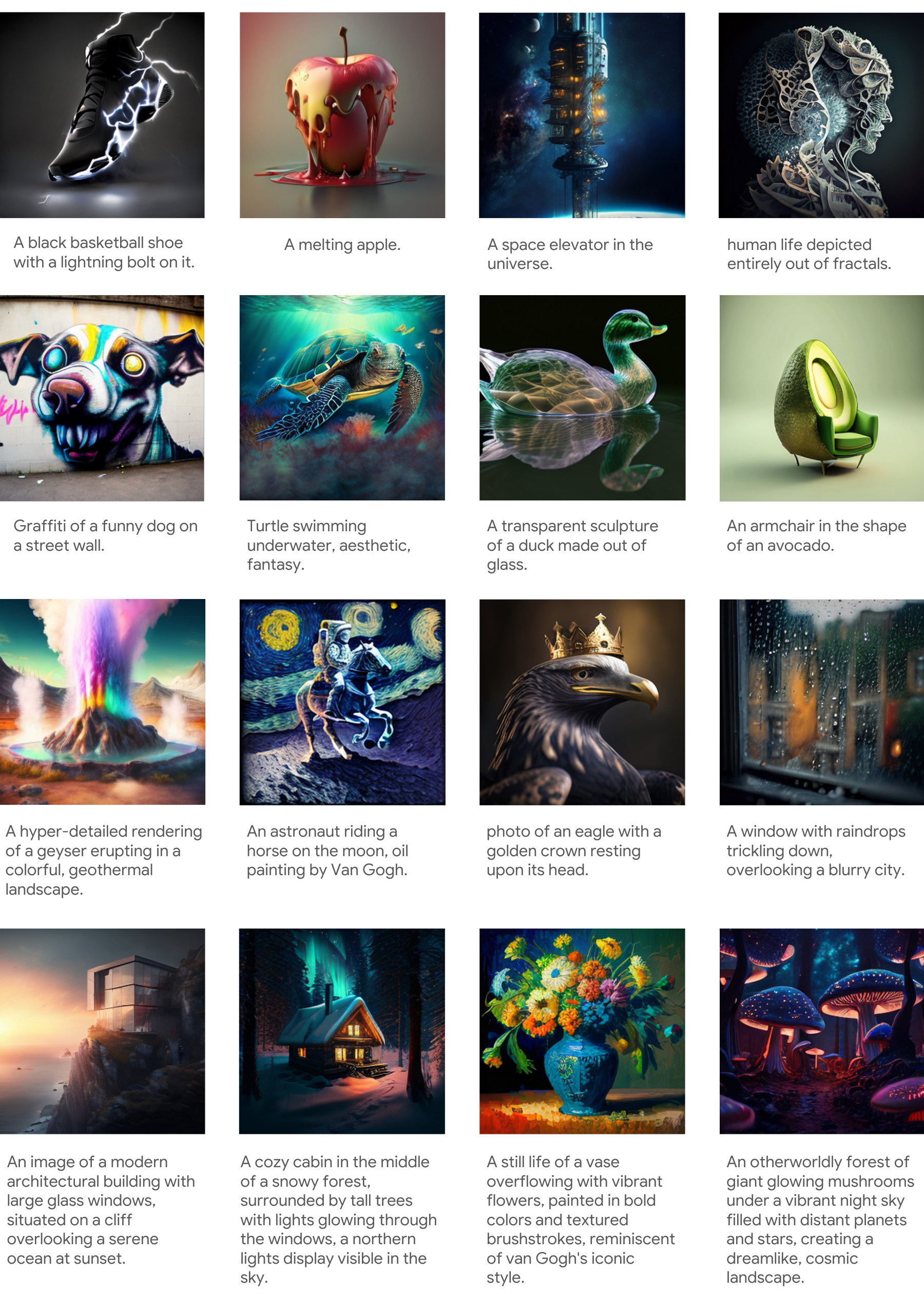}
\vspace*{-7mm}
\centering
\caption{
\small Images generated from \name~autoregressive model after aesthetic fine-tuning. 
}
\label{fig:sft}
\end{figure}
The experimental results are presented in Table~\ref{tab:backbone_llm}. Here we used $\lambda=0.005$ for all models. The results demonstrate that employing a stronger LLM is crucial for generating images with higher fidelity and quality. Gemma-2 achieves significantly lower FID scores compared to Gemma-1, highlighting that even though LLM pre-training is unimodal, without exposure to visual data, using a better LLM as a backbone is important for unlocking improved visual quality in a unified model training setup. 
The image understanding performance also improved slightly when using stronger LLM, which is a trend also noted in PaliGemma2. We show the detailed comparison on each downstream visual understanding benchmark in Figure~\ref{fig:ds_gemma}.

\textbf{Training with Random Order Helps Generation But Not Understanding.}
Images inherently possess 2D patterns. As demonstrated in Fluid, raster-order training can be problematic, potentially leading to collapse and artifacts characterized by disappearing patterns.
Approaches such as RAR~\cite{yu2024randomized} and RandAR~\cite{pang2024randar} propose training image generation AR models with random-order training, which can improve ImageNet FID and result in better visual quality.
\begin{table}[t]
\caption{\small Performance comparison of image generation and understanding of \name~trained with different image generation order. FID and CIDEr is measured on MS-COCO.}
\label{tab:generation_order}
\centering
\resizebox{0.82\textwidth}{!}{
\begin{tabular}{ccc|ccc}
\toprule
\multirow{2}{*}{Generation Order} & \multicolumn{2}{c|}{Generation} & \multicolumn{3}{c}{Understanding} \\ \cmidrule{2-6} 
                 & COCO FID $\downarrow$  & GenEval $\uparrow$  & COCO CIDEr $\uparrow$ &  Cap Avg $\uparrow$ &  QA Avg $\uparrow$ \\ 
                 \midrule
Raster          & 8.28     & 0.59    & \bf 45.57 & \bf 116.55 & 61.81 \\
Random          & \bf 7.20      & 0.59    & 40.91 & 116.13 & \bf 62.10 \\ 
\bottomrule
\end{tabular}
}
\end{table}

Here we study the effect of different visual generation orders during training within our unified framework setup. We compare the performance between random-order and raster-order training, both with Gemma-2 2B as backbone LLM.
The results presented in Table~\ref{tab:generation_order} indicate that for per-token image generation within a unified framework, raster-order training continues to underperform compared to random-order generation. Incorporating random-order during training could ensure the generated images are of high quality. However, it does not necessarily improve the visual understanding performance, where raster-order achieves better performance on MS-COCO CIDEr and downstream captioning task average (Cap Avg).

\subsection{More Generation Capabilities}

We also verify the transferability of the trained model to various downstream generation tasks.

\textbf{Aesthetic Fine-Tuning.}
To enhance the visual quality and aesthetic appeal of the generated images, we perform aesthetic fine-tuning on %
a publicly available dataset. The results are shown in Figure~\ref{fig:sft}.

\textbf{Image Editing Task.}
Since our unified framework is trained with multimodal inputs, it can naturally extend to image editing tasks that involve both image and text prompt inputs. We fine-tune the 2B \name~model with 4M image editing pairs from HQEdit~\cite{hui2024hq} and UltraEdit~\cite{zhao2024ultraedit}. In Figure~\ref{fig:editing},
we apply the fine-tuned model to the input images and editing prompts from a public available benchmark.
Although preliminary, the experiments show that \name~ is able to adapt and generalize to tasks that involve interleaved data modalities.

\section{Conclusion}
In this paper, we presented \name, a pure autoregressive framework for joint visual generation and understanding, utilizing continuous visual tokens. 
We identified an inherent trade-off between the visual generation and understanding tasks, but the two tasks can benefit each other with tuned training recipes.
Careful choice of the loss balance between the two tasks allows a single unified model to achieve performance comparable to or exceeding single-task baselines.
We conducted investigation of key design choices for \name~training, revealing the critical importance of employing strong backbone LLM and random-order generation to unlock high-quality visual generation capabilities.
We believe that this work encourages future research into the exploration of continuous visual tokens for unified vision-language model training, paving the way for more efficient and powerful autoregressive multimodal systems.

\textbf{Acknowledgements.} We are grateful to Alex Rizkowsky and Amy Shen for their support in securing computational resources. We also wish to thank Charles Herrmann, Junhwa Hur, Shangbang Long, André Susano Pinto, Srinivas Kaza, David Salesin, and the VisCam team for their insightful discussions and constructive feedback, which greatly improved this work.

\clearpage

\bibliographystyle{plain}
\bibliography{unifluid}

\end{document}